\def\year{2021}\relax
\documentclass[letterpaper]{article} 
\usepackage{aaai21}  
\usepackage{times}  
\usepackage{helvet} 
\usepackage{courier}  
\usepackage[hyphens]{url}  
\usepackage{graphicx} 
\usepackage{natbib}  
\usepackage{caption} 
\usepackage{booktabs} 
\usepackage{amsthm}
\usepackage[]{algorithm2e}
\usepackage{etoolbox}
\usepackage{array}
\usepackage{natbib}
\setcitestyle{authoryear}
\usepackage{amsmath,amssymb,amsfonts}
\usepackage{algorithmic}
\usepackage{textcomp}
\usepackage{xcolor}
\usepackage[font=scriptsize]{caption}
\usepackage{caption}
\usepackage{subcaption}
\usepackage{array}
\usepackage{pgfplots}
\usepackage{multicol}
\usepackage{subcaption}
\usepackage{dsfont}
\DeclareMathOperator*{\argmax}{arg\,max} 
\makeatletter

\def\addlegendimage{\pgfplots@addlegendimage}
\makeatother

\title{Towards Robust Deep Learning \\With Ensemble Networks and Noisy Layers}
\author {
     Yuting Liang\textsuperscript{\rm 1}
     Reza Samavi\textsuperscript{\rm 2,3} \\
 }
 \affiliations {
     \textsuperscript{\rm 1}McMaster University, 
     \textsuperscript{\rm 2}Ryerson University,
     \textsuperscript{\rm 3}Vector Institute \\
     liangy42@mcmaster.ca, samavi@ryerson.ca
 }
\begin{document}

\maketitle

\begin{abstract}
In this paper we provide an approach for deep learning that protects against adversarial examples in image classification-type networks. The approach relies on two mechanisms:1) a mechanism that increases robustness at the expense of accuracy, and, 2) a mechanism that improves accuracy but does not always increase robustness. We show that an approach combining the two mechanisms can provide protection against adversarial examples while retaining accuracy. We formulate potential attacks on our approach with experimental results to demonstrate its effectiveness. We also provide a robustness guarantee for our approach along with an interpretation for the guarantee.
\end{abstract}

\section{Introduction}
\label{Introduction}
Deep neural networks (DNN) are increasingly being adapted to perform a wide range of tasks from navigation and personal recommendation systems for consumer use to a larger scale decision making systems such as speech recognition and computer vision. However, application of DNN in safety critical systems is hampered by its vulnerability to \emph {adversarial examples}, where an adversary uses carefully crafted small amounts of perturbations to force the DNN to make erroneous classifications in the inference phase. Since the first published attack by \cite{szegedy2013intriguing} there has been many attempts to eliminate or otherwise substantially increase the cost for crafting such attacks \cite{goodfellow2014explaining,kurakin2016adversarial,papernot2016distillation}. However, it was shown that many such defences were later defeated by more advanced attacks \cite{carlini2017towards}. Since then, there has been a shift in research focus from \emph{empirical} defences to measuring DNN's robustness in terms of its resistance to adversarial examples \citep{bastani2016measuring,weng2018evaluating,hendrycks2019benchmarking}, and more recently to finding solutions that offer robustness guarantees \citep{lecuyer2019certified,cohen2019certified}.
%
    
In this paper, we provide an approach toward increasing robustness of DNN against adversarial examples. Our intuition for creating such a defence relies on the idea of training an ensemble of networks and using the aggregate outputs of the networks to decide on a final output. A similar voting mechanism proved successful for protecting data privacy via differential privacy in PATE~\citep{papernot2016semi}, where the training set containing sensitive data is partitioned into disjoint sets, each is used for training a \emph {teacher} network. The mechanism of PATE relies on partitioning and keeping the sensitive data secret to provide data privacy. 
In our case, the intent of using an ensemble is to minimize the probability of a successful attack by increasing the amount of effort, and potentially increasing the amount of perturbation to an image so the attack becomes noticeable (i.e. fails). We provide empirical results to demonstrate its effectiveness and an analysis into its robustness properties.
\section {Background}
\subsection {L-BFGS}
The problem of creating an adversarial example is to perturb some pixels on an image such that the classifier will label the perturbed image differently from the original image's correct label. These adversarial examples are generally categorized as being \emph {targeted}, where the adversary carefully adds perturbations to cause the classifier to output  a specific incorrect label; or \emph {untargeted}, in which the adversary's goal is to cause the classifier to output any incorrect label. Naturally, one can think of both types of adversarial examples as targeted with a set of target labels. In the former case there is only one label in the target set, whereas in the latter case the target set consists of all incorrect labels. Intuitively if one keeps adding random perturbations to an image, eventually even the most sophisticated classifier (including humans) will start to produce incorrect labels. Moreover, such perturbed images likely will not appear genuine. Thus, for an adversarial example to be considered successful, the amount of perturbations to be made to the original image must be minimal. Formally, we can define the problem of finding adversarial examples as
{\small
\begin{equation} \label{formal}
\min_{x' \in D} ||x'-x||   s.t. \: F(x') = t
\end{equation}
}
for some classifier $F(\cdot)$,  where $t$ is the target label, $x$ is the original input image. $D$ is the domain of all images which are usually represented as a set of multi-dimensional vectors, where the permissible values fall within a bounded range, with some distance metric $||\cdot||$. This definition is used in various literature up to notational differences.

The first published attack by \cite{szegedy2013intriguing} on neural networks arose by solving the above problem using a box-constrained L-BFGS method, where $D=[0,1]^m \subseteq \mathbb{R}$. In the L-BFGS example, the above problem is first transformed from a constrained optimization problem to an unconstrained optimization problem (up to the box constraints) using a penalty method, where the penalty function is the loss function applied to $x'$ and the target $t$. In other words, for constant $c > 0$, the above problem becomes:
{\small
\begin{equation} \label{L-BFGS}
\min_{x' \in D} c\cdot||x'-x||+L_F(x',t).
\end{equation}
}
The loss function is appealing as the penalty function because it's non-negative and captures the distance between $F(x')$ and $t$ with the same metric used in the training of the network. Moreover, the loss function is zero exactly when the original constraint $F(x') = t$ is satisfied. Thus an optimal solution to the original problem will also be an optimal solution to the transformed problem. 

L-BFGS belongs to the class of line search methods which are commonly used for solving optimization problems over a compact domain. A line search method is an iterative scheme, and typically involves repeatedly identifying a search direction and moving the feasible point along the search direction. The scheme will converge to a global optimum if the formulation satisfies certain conditions. In the problem above, if the loss function is sufficiently differentiable, then a global optimum always exists. In other words, an adversarial example is guaranteed to exist; however the perturbation $||x'-x||$ is not necessarily small.

\subsection {On Gradient-Based Attacks}
\label{gradient-based}
Following the invention of the L-BFGS adversarial examples, other methods such as \cite{goodfellow2014explaining, kurakin2016adversarial, papernot2016limitations} soon emerged which further accentuated the presence of significant vulnerabilities in classification-type networks. Shortly after these inventions, a technique for training classification-type networks, known as Defensive Distillation \cite{papernot2016distillation} was devised, which employs the idea of using a \emph{teacher} network to train a second network (\emph{student}).

Defensive Distillation was soon defeated by the Carlini-Wagner attack \citep{carlini2017towards}. The success of Carlini \& Wagner demonstrated the important fact that the prior success of Defensive Distillation was due to the different temperature constants used between the \emph{teacher} and \emph{student} network, as well as the assumption that attacks would be crafted based on the original temperature (same as used for the teacher). In particular, Carlini \& Wagner highlighted a common characteristic in the prior attacks that Defensive Distillation successfully defended against: that the attacks depend on the the gradient of the network, either used as a multiplicative term in the search direction or to determine the amount of perturbations to add to specific pixels. 

The softmax function is usually used for classification-type networks at the second last layer to normalize outputs into discrete probabilities, i.e., for network $F$, we can write $F:= \sigma \circ G$ for softmax function $\sigma$, where $G$ is the composition of all previous layers. As such, the gradient of the network contains the derivative of the softmax function as a multiplicative term:
{\small
\begin{equation} \label{partial_F}
    \frac{\partial F}{\partial x_k} = \frac{\partial F}{\partial z^j}(\frac{\partial z^j}{\partial x})
    = \frac{\partial \sigma(z^j)}{\partial z^j}(\frac{\partial z^j}{\partial x})
    = \frac{1}{T}z^j(\frac{\partial z^j}{\partial x_k}),
 \end{equation}
 }
for input pixel $x_k$, where the softmax function $\sigma(\cdot)$ at the second last layer for the $j^{th}$ classification is
{\small
  \begin{equation} \label{softmax_T}
   \sigma(z^j) = \frac{e^{z^j/T}}{\sum_i e^{z^i/T}},
 \end{equation}
 }
and $z^j := [G(x_k)]_j$ is the $j^{th}$ component of the output from the previous layers.
In Defensive Distillation, a large temperature $T$ is used for the student network, but the attacks are assumed to be crafted with $T=1$ in the softmax function. The larger $T$ reduces the gradient by $T$,
essentially disabling the search to advance toward an optimal solution. Thus, it becomes clear that a \emph{tampered} gradient can impair the advance of any adversarial attacks crafted using the gradient of the original network function. Noting this observation, in order to remove the dependency on the gradient of the network, Carlini \& Wagner re-formulated the problem in (\ref{L-BFGS}). 
In particular, they introduced different penalty functions which do not depend on the original network output; instead they advertised the use of penalty functions that depend on the output at the second last layer (i.e. the \emph {logits}). Thus, a more general formulation for finding adversarial examples is, for penalty function $L(\cdot)$ not necessarily equal to the original loss function:
{\small
\begin{equation} \label{adv_problem}
    \min_{x' \in D} c\cdot||x'-x||+L(x',t).
 \end{equation}
 }
Since the softmax function applied at the last layer is monotonic, the final output is already decided at the second last layer, thus equivalent penalty functions to that in (\ref{L-BFGS}) can be created which do not depend directly on the original output. Moreover, the logits do not depend on the temperature constant and thus are not impacted by larger temperature constant used in Defensive Distillation.

\section{The Model}
%
%

\subsection{Noisy Logits}
Since Carlini-Wagner attacks require access to the logits in the solution to (\ref{adv_problem}), we can obscure the search for solution by adding random noise to the logits. In fact, we can obscure any iterative scheme by adding random noise to any layer whose output is needed for crafting an attack. Note that if we add random noise directly to the original logits, an adversary might recover the original logits by making multiple queries and averaging the resulting \emph{noisy} logits. Instead, we apply random noise at query time to the input, then respond to the query with the logit of the perturbed input. Also, since the softmax function is monotonic, we must ensure the final output is a result of the \emph{noisy} logit, otherwise the genuine logit can be recovered by applying the inverse of the softmax function to the result at the output layer. Let $z_0$ be an input with $F(z_0)$ its output from the network $F$. Moreover, suppose $F$ has $n$ layers besides the input layer and for $1 \leq i \leq n$
{\small
\begin{equation}
z_i := F_i \circ F_{i-1} \circ \cdots \circ F_1(z_0),
\end{equation}
}
where $\circ$ denotes composition and
{\small
\begin{equation}
F(z_0) := z_n = F_n \circ F_{n-1} \circ \cdots \circ F_1(z_0).
\end{equation}
}
Then at the output layer $i$, the Noisy Logit mechanism will produce output
{\small
\begin{equation}
F_i(z_{i-1}') = F_i \circ F_{i-1} \circ \cdots \circ F_1(z_0'),
\end{equation}
}
where $z_0' = z_0 + g(\vec{a})$ and $g(\vec{a})$ is a random noise function with parameter(s) $\vec{a}$.
Note that by this procedure, naturally we can respond to queries at any layer with a noisy output, thus preventing an adversary from trying to reconstruct a genuine output at any layer (including the logits) by making queries to the network.

We remark the important distinction between noise injection at training time vs. at query time. Training a network with noise to over-fit the predictions to some neighbourhood of the input can induce vulnerability to \emph{invariance-based} adversarial examples \citep{jacobsen2019exploiting}, where the adversary inserts enough changes to an input such that its label should be changed but it's still predicted as the original label by the model. 
We opt to inject noise at query time so as to obscure iterative schemes for adversarial example constructions. 

Noise injection at query time is also used by the authors in \cite{yang2019me}, but this noise is accompanied by noise injection at training time. In their approach, the goal of noise injection is to mask the adversarial perturbations. The masked inputs are then reconstructed to reveal close approximations to the original inputs where the perturbations have been effectively smoothed out. The network is then trained and tested on the reconstructed inputs. In contrast, we do not attempt to remove the adversarial perturbations nor the noise added. Inevitably, this added noise might cause a network to lose accuracy. To compensate for the potential loss in accuracy, we make use of ensemble networks as described below.

\subsection{Ensemble Voting}
The purpose of having an ensemble of networks is two-fold: 1) to provide resilience in the \emph{combined} network when some of the networks are under attack; 2) to improve accuracy in the \emph{combined} network over the individual networks.

If we require that each individual network must successfully classify the input image, then assuming independence of success probabilities across the networks, the probability of simultaneous success across the networks is the product of the success probability of each network, which might be less than a desirable level of accuracy if the total number of networks is large since we are multiplying a series of numbers less than 1. However, if we only require success in the largest subset of the networks, then since there are many possible permutation of subsets when the number of networks is large, the success probability of the combined network as an aggregate might be much better than that of each individual network. Let $S:=\{F^1,F^2,...,F^m\}$ be a collection of $m$ networks, let $\delta(S)$ be the set of all partitions of $S$. For each partition $h\in\delta(S)$, let $L_h$ denote the largest subset in $h$. Then, the probability of success by voting is $\sum_{h\in\delta(S)} P(L_h)$, 
where $P(L_h)$ is the probability of simultaneous success in $L_h$. Note that when $m$ is large, the number of possible partitions is large which means the success probability by voting can be high.

We note that although Carlini-Wagner attacks are able to defeat a network trained with any temperature constant, an attack crafted for a network trained with one temperature constant might not work on another trained with a different temperature, due to differences in the trained parameters. Thus we propose a mechanism where we train an ensemble of networks, each trained with a different temperature constant, where we respond to queries using the aggregate outputs of the ensemble of networks.

\subsection{Rank Verification}
In a system that employs voting as a mechanism, the final outputs can be somewhat noisy depending on the number of participants, whereby an addition of another participant can sometimes alter the final outcome. In our context, if two classes have similar levels of votes in the ensemble, the networks see the input as bearing resemblance to both classes, which could be an indication that the input could be easily compromised (if not already compromised). In this case, it makes sense to \emph{abstain} from making a prediction or warn the user of the potential risk. This intuition is supported by the findings in \cite{dogus2018intriguing}, where the authors demonstrated that adversarial accuracy is dictated by the distribution of differences between the values of the logits corresponding to the most likely and the second most likely classes. In particular, the success of many adversarial examples is due to such differences being small. \\
We follow the Rank Verification method proposed in \citep{hung2019rank}, which tests the hypothesis that the top two candidates are equally likely to be the "winner" in a voting system. Let $n_A$ and $n_B$ correspond to the vote counts of the top two classes $y_A$ and $y_B$, respectively, predicted by the ensemble for some given input $x$, where $n_A \geq n_B$. Let $p=\textit{BinomTest}(n_A,n_A+n_B,0.5)$
be the p-value obtained from the hypothesis test for $n_A$ observations of $y_A$ over $n_A+n_B$ trials, with hypothesized success probability equal to 0.5. If $p<\alpha_{RV}$, we reject the hypothesis at some significance level $\alpha_{RV}$. i.e. $y_A$ is most statistically likely to be the winner among the outputs from the ensemble. If we cannot reject the hypothesis, we abstain from making a prediction at $x$ or issue a warning.

%
%

\section{Robustness Guarantee}
It is important to understand how much perturbation the model is able to withstand without changing its prediction, as it gives us confidence about the model's {\emph{robustness}} to make predictions under adversarial attacks. We say a model is robust to perturbations of a certain size when it does not alter its original prediction on an input even after perturbations up to a certain size have been added to the input. Formally, let $F:D\rightarrow\mathcal{Y}$ denote a classifier, let $x \in D$ be any input. We say $F$ is robust at $x$ up to perturbations of size $R$ (under some norm function $||\cdot||$), if for any $\delta \in \Re^m$ such that $||\delta||\leq R$ and $x+\delta \in D$, we have
{\small
\begin{equation}
F(x+\delta)=F(x).
\end{equation}
}
In this section, we provide a Robustness Guarantee for our approach combining Ensemble Voting and Noisy Logit. We follow the analysis in \citep{cohen2019certified} and adapt their guarantee and certification procedure to our approach. Let $F^*$ denote the classifier  whose output is given by the most frequent output among an ensemble of networks $\{F^l:l=1,...,m\}$, or 
\begin{equation} \label{eqn:ensemble}
F^*(x)=\argmax_{y\in\mathcal{Y}} \sum_{l=1}^{m} \mathds{1}_{F^l(x)=y}.
\end{equation}
Let $\varepsilon ~ N(0,\sigma^2I)$, define
\begin{equation} \label{eqn:noisy_ensemble}
g(x)=F^*(x+\varepsilon).
\end{equation}
\textbf{Theorem 1} ($L^2$ norm): Let $F^*$ be as defined in equation (\ref{eqn:ensemble}) and $g$ be as defined in equation (\ref{eqn:noisy_ensemble}). If there exists $y_A \in \mathcal{Y}$ and $\underline{p_A}, \overline{p_B} \in [0,1]$ such that
\begin{equation}
\mathds{P}(F^*(x+\varepsilon)=y_A)\geq\underline{p_A}\geq\overline{p_B}\geq\max_{y\neq y_A} \mathds{P}(F^*(x+\varepsilon)=y),
\end{equation}
then $g(x+\delta)=y_A$ for all $||\delta||_2 \leq R$, where
\begin{equation}
R=\frac{\sigma}{2}(\Phi^{-1}(\underline{p_A})-\Phi^{-1}(\overline{p_B})).
\end{equation}
The proof of this result is almost identical to that in \citep{cohen2019certified}; in particular, in the proof of \citep{cohen2019certified} their network can be replaced by our ensemble $F^*$, and their smoothed classifier can be replaced by our noisy ensemble $g$. Note that a similar result for the $L^1$ norm also exists and was provided by \citep{teng2020ell}. Robustness in $L^1$ norm is naturally induced by Laplace noise just as robustness in $L^2$ norm is naturally induced by Gaussian noise.

\subsection{Certification}
It is difficult to compute exactly the probability of $F^*$ at any input since we must consider the joint distribution of the ensemble, which requires computing their correlations, and the immense range of possible permutations of subsets among the ensemble. Computing the probability on a noisy input ($x+\varepsilon$) is an even more challenging task. We can, however, approximate the distribution of $F^*(x+\varepsilon)$ for a given input $x$ using Monte Carlo simulations, from which we can then compute the quantities $\underline{p_A}$ approximately as the lower confidence interval at a desired significance level $\alpha$. Specifically, suppose the computed output is $\hat{y_A}=F^*(x+\hat{\varepsilon})$; we can think of the output distribution as consisting of two outcomes: $\{\hat{y_A}\}$ and $\mathcal{Y} \setminus \{\hat{y_A}\}$. Thus, $\underline{p_A}$ can be viewed as a lower bound for the probability of success in binomial distribution. Once we obtain $\underline{p_A}$, we can approximate $\overline{p_B}$ as $1-\underline{p_A}$ so long as $\underline{p_A} > 1/2$. This gives $R=\frac{\sigma}{2}(\Phi^{-1}(\underline{p_A})-\Phi^{-1}(1-\underline{p_A}))=\sigma\Phi^{-1}(\underline{p_A})$.
\begin{algorithm} [h]
\mbox{Certification Procedure}\\
 \LinesNumbered
 \textbf{Input:}  $x$, $\sigma$, $n$, $\alpha$, $F^*$\; 
 \textbf{Output:}  $y_A$, $p$, $R$ \; 
$\hat{\varepsilon}$ $\leftarrow$ $N(0,\sigma^2)$ draw a noise sample\;
$\hat{y_A}$, $\hat{y_B}$ $\gets F^*(x+\hat{\varepsilon})$ top 2 classifications\;
$\hat{n_A}$, $\hat{n_B}$ $\gets$ vote counts corresponding to $\hat{y_A}$, $\hat{y_B}$\;
$p \gets$ \textit{BinomTest}$(n_A,n_A+n_B,0.5)$\;
$A \gets \emptyset$\;
\For{$i:=1$ to $n$}{  
$\varepsilon_i$ $\leftarrow$ $N(0,\sigma^2)$ draw noise sample\;
$A[i] \gets F^*(x+\varepsilon_i)$\;
$n_A \gets$ counts of $\hat{y_A}$ in $A$\;
$\underline{p_A} \gets$ \textit{ConfIntLower}$(n_A,n,\alpha)$\;
}
if $\underline{p_A} \leq 0.5$ then $R = 0$;
else $R = \sigma\phi^{-1}(\underline{p_A})$ \;
\Return($\hat{y_A}$, $p$), $R$ 
\label{certify}
\end{algorithm}
Note that strictly speaking $p$ is not required for the Certification Procedure, however it provides indication for the confidence of the prediction $\hat{y_A}$. In particular, when it's large (relative to some desired significance level) we interpret that the prediction might not be reliable or that the input might be easily compromised. Indeed, we observed that a larger value of $p$ would often lead to a smaller value for $\underline{p_A}$.

\section{Experimental Evaluation}
\label{testing_results}
In this section we assess the effectiveness of our approach by formulating two types of potential attacks on our model. In the first type we consider an attack crafted for a randomly chosen network in the ensemble. Since the model architectures are so similar across the networks, transferability is possible where networks other than the chosen one might still incorrectly classify. In the second type of attacks we consider superimpositions of adversarial examples to examine whether this could further increase transferability. For a chosen subset of the networks, each with a corresponding adversarial example, one could reasonably suspect that other networks beyond the chosen subset could incorrect classify as the superimposition could have captured perturbations that are commonly effective on many other networks\footnote {Additional experiments plus the source code for the experiments can be found at: \url{https://tinyurl.com/y4qkabd4} .}.

\subsection{Test Setup}
\label{test_setup}
We conducted experiments on two datasets, MNIST \cite{lecun1998} and CIFAR10 \cite{krizhevsky2009learning}. For the following tests, the architecture of each network is the same as the one in \cite{carlini2017towards}, which we provide in Table~\ref{tab:ensemble_architecture}. We first trained an ensemble of networks $F^l$, each with temperature $T_l, l=1,2,...,m$. We used $T_l=10\cdot l$, $m=50$.
In all the experiments, the $L^2$ norm is used in all places where a norm is needed, including the $L^2$ version of the Carlini-Wagner attack. For the models with Noisy Logit, we employ a 
Gaussian noise function $g(0,\sigma^2)$ with $\sigma=0.5$ for MNIST and $\sigma=0.03$ for CIFAR10. We experimented with different values for the scale parameter, and found these values provide sufficient noise without losing too much accuracy. 

\begin{table}[t]
\centering
\begin{subtable}{.8\linewidth}
\subcaption{Model architectures} 
\vspace {-.2cm}
\centering
\resizebox{0.88\textwidth}{!}{
\begin{tabular}{|l|l|l|}
\hline
\textbf{Layer}         & \textbf{MNIST} & \textbf{CIFAR10} \\ \hline
Convolution + ReLU     & $3\times3\times32$ & $3\times3\times64$ \\ \hline
Convolution + ReLU     & $3\times3\times32$ & $3\times3\times64$ \\ \hline
Max Pooling            & $2\times2$    & $2\times2$    \\ \hline
Convolution + ReLU     & $3\times3\times32$ & $3\times3\times64$ \\ \hline
Convolution + ReLU     & $3\times3\times32$ & $3\times3\times64$  \\ \hline
Max Pooling            & $2\times2$    & $2\times2$    \\ \hline
Fully Connected + ReLU & $200$    & $256$    \\ \hline
Fully Connected + ReLU & $200$    & $256$    \\ \hline
Softmax                & $10$     & $10$    \\ \hline
\end{tabular}%
}
\label{tab:ensemble_architecture}
\end{subtable}
\quad
\begin{subtable}{.8\linewidth} 
\centering
\vspace{.2cm}
\subcaption{Parameters used}
\vspace {-.2cm}
\resizebox{0.86\textwidth}{!}{%
\begin{tabular}{|l|l|l|}
\hline
\textbf{Parameter}              & \textbf{MNIST} & \textbf{CIFAR10} \\ \hline
Learning Rate                   & $0.01$           & $0.01$           \\ \hline
Decay                           & $1.00e-06$       & $1.00e-06$       \\ \hline
Momentum                        & $0.9$            & $0.9 $           \\ \hline
Dropout                         & $0.5$            & $0.5$            \\ \hline
Batch Size                      & $128$            & $128$            \\ \hline
Partitioned Training Set        & Yes              & No             \\ \hline
Training Set Size (Per Network) & $1100$           & $45000$          \\ \hline
Validation Set Size             & $5000$           & $5000$           \\ \hline
Epochs                          & $3000$           & $150$            \\ \hline
Gaussian Noise Sigma            & $0.5$            & $0.03$           \\ \hline
\end{tabular}%
}
\label{tab:ensemble_parameter}
\end{subtable}
\vspace {-.2cm}
\caption{Setup for the ensemble of networks} 
\end{table}
\noindent
\textbf{MNIST} We first partitioned the original dataset into 50 training subsets (1100 samples each) and one validation set (5000 samples). We trained 50 teachers individually on the partitioned subsets, each was trained with a different temperature constant for 3000 epochs. The average validation accuracy among the individual networks was ~94.03\%.

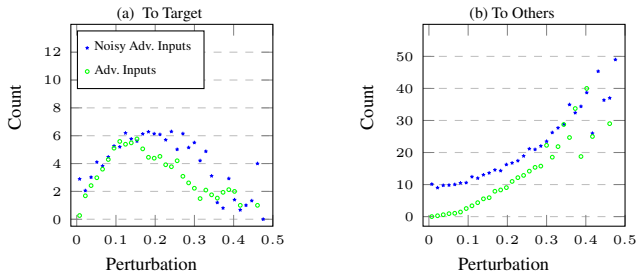
\begin{figure}[t]
\centering
\begin{multicols}{2}
\begin{subfigure}{0.24\textwidth}
\subcaption{\label{fig:count_changed_target_mnist} To Target }
\vspace {-.2cm}
\begin{tikzpicture}
\begin{axis}[
    xlabel={Perturbation},
    ylabel={Count},
   xlabel style={at={(0.4,1.5ex)}},
    ylabel style={at={(0.18,10ex)}},
    xmin=-0.015, xmax=0.5,
    ymin=-.5, ymax=14,
    xtick={0,.1,.2,.3,.4,.5},
    ytick={0,2,4,6,8,10,12},
    mark size=0.7pt,
    ymajorgrids=true,
    grid style=dashed,
    legend style={font=\fontsize{5}{3}\selectfont},
    legend pos=north west,
    legend cell align={left},
    label style={font=\fontsize{7}{3}\selectfont},
    tick label style={font=\fontsize{4}{3}\selectfont},
    width=\textwidth,
    height=\textwidth,   
]
 
\addplot[
only marks,
    color=blue,
    mark=star,
    ]
file [skip first] {Plots/mnist_normal/Figure_flip2target_count.png_1.txt};
    \legend{Noisy Adv. Inputs}
    
\addplot[
only marks,
    color=green,
    mark=o,
    only marks,
    ]
file [skip first] {Plots/mnist_normal/Figure_flip2target_count.png_0.txt};
    \addlegendentry{Adv. Inputs}    

\end{axis}
\end{tikzpicture}
\end{subfigure}
\columnbreak

\begin{subfigure}{0.24\textwidth}
\subcaption{\label{fig:count_changed_other_mnist}To Others}
\vspace {-.2cm}
\begin{tikzpicture}
\begin{axis}[
    xlabel={Perturbation},
    ylabel={Count},
   xlabel style={at={(0.4,1.5ex)}},
    ylabel style={at={(0.18,10ex)}},
    xmin=-0.015, xmax=0.5,
    ymin=-3, ymax=60,
    xtick={0,.1,.2,.3,.4,.5},
    ytick={0,10,20,30,40,50},
    mark size=0.7pt,
    ymajorgrids=true,
    grid style=dashed,
    legend style={font=\fontsize{5}{3}\selectfont},
    legend cell align={left},
    label style={font=\fontsize{7}{3}\selectfont},
    tick label style={font=\fontsize{4}{3}\selectfont},
    width=\textwidth,
    height=\textwidth,   
]
 
\addplot[
    color=blue,
    mark=star,
    only marks,
    ]
file [skip first] {Plots/mnist_normal/Figure_flip2other_count.png_1.txt};
    
\addplot[
    color=green,
    mark=o,
    only marks,
    ]
file [skip first] {Plots/mnist_normal/Figure_flip2other_count.png_0.txt};
    
\end{axis}
\end{tikzpicture}
\end{subfigure}
\end{multicols}
\vspace {-.6cm}
\caption{\label{fig:changed_mnist_single} MNIST: Counts of Single Networks Changed}
\end{figure}

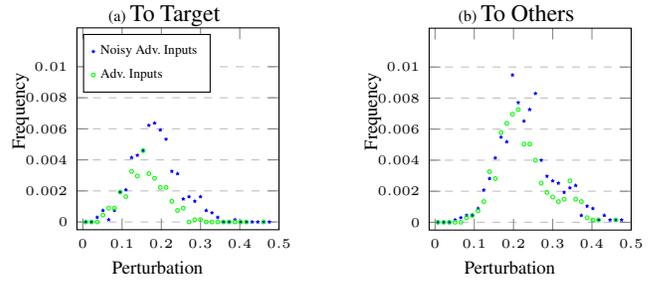
\begin{figure}[t]
\begin{multicols}{2}
\centering
\begin{subfigure}{0.24\textwidth}
\subcaption{\label{fig:freq_changed_target_mnist}{\footnotesize To Target}}
\vspace {-.2cm}
\begin{tikzpicture}
\begin{axis}[
    xlabel={Perturbation},
    ylabel={Frequency},
   xlabel style={at={(0.4,1.5ex)}},
    ylabel style={at={(0.18,10ex)}},
    xmin=-0.015, xmax=0.5,
    ymin=-0.0005, ymax=0.0125,
    xtick={0,.1,.2,.3,.4,.5},
    ytick={0.0, 0.002, 0.004, 0.006, 0.008,0.010},
    yticklabel style={/pgf/number format/fixed, /pgf/number format/precision=3},
    scaled y ticks=false,
    mark size=0.7pt,
    ymajorgrids=true,
    grid style=dashed,
    legend style={font=\fontsize{5}{3}\selectfont},
    legend pos=north west,
    legend cell align={left},
    label style={font=\fontsize{7}{3}\selectfont},
    tick label style={font=\fontsize{4}{3}\selectfont},
    width=\textwidth,
    height=\textwidth,   
]
 
\addplot[
only marks,
    color=blue,
    mark=star,
    ]
file [skip first] {Plots/mnist_normal/Figure_adv2target_freq.png_1.txt};
    \legend{Noisy Adv. Inputs}
    
\addplot[
only marks,
    color=green,
    mark=o,
    only marks,
    ]
file [skip first] {Plots/mnist_normal/Figure_adv2target_freq.png_0.txt};
    \addlegendentry{Adv. Inputs}    

\end{axis}
\end{tikzpicture}
\end{subfigure}
\columnbreak

\begin{subfigure}{0.24\textwidth}
\subcaption{\label{fig:freq_changed_other_mnist}{\footnotesize To Others}}
\vspace {-.2cm}
\begin{tikzpicture}
\begin{axis}[
    xlabel={Perturbation},
    ylabel={Frequency},
   xlabel style={at={(0.4,1.5ex)}},
    ylabel style={at={(0.18,10ex)}},
    xmin=-0.015, xmax=0.5,
    ymin=-.0005, ymax=0.0125,
    xtick={0,.1,.2,.3,.4,.5},
    ytick={0.0, 0.002, 0.004, 0.006, 0.008, 0.010},
    yticklabel style={/pgf/number format/fixed, /pgf/number format/precision=3},
    scaled y ticks=false,
    mark size=0.7pt,
    ymajorgrids=true,
    grid style=dashed,
    legend style={font=\fontsize{5}{3}\selectfont},
    legend cell align={left},
    label style={font=\fontsize{7}{3}\selectfont},
    tick label style={font=\fontsize{4}{3}\selectfont},
    width=\textwidth,
    height=\textwidth,   
]
 
\addplot[
    color=blue,
    mark=star,
    only marks,
    ]
file [skip first] {Plots/mnist_normal/Figure_adv2other_freq.png_1.txt};
    
\addplot[
    color=green,
    mark=o,
    only marks,
    ]
file [skip first] {Plots/mnist_normal/Figure_adv2other_freq.png_0.txt};
    
\end{axis}
\end{tikzpicture}
\end{subfigure}

\end{multicols}

%

\vspace {-.6cm}
\caption{\label{fig:changed_mnist_aggregate}  MNIST: Ensemble Aggregate Outputs Changed}

\end{figure}

\noindent
\textbf{CIFAR10} In the setup for CIFAR10 we used a single training set (45000 samples) and one
validation set (5000 samples). We do not use the networks trained on partitioned datasets for testing on CIFAR10, because we observed that a small training dataset resulted in low accuracy for CIFAR10. We trained 50 teachers individually on the same training set, each was trained with a different temperature constant for 150 epochs. The average validation accuracy among the individual networks was ~72.42\%.

   \begin{table}[h]
 \centering
\begin{subtable}{1.0\linewidth} 
\centering
\subcaption{Accuracy}
\vspace{-0.2cm} 
    \resizebox{0.95\textwidth}{!}{%
\begin{tabular}{@{}lll@{}}
\toprule
\textbf{Model}                  & \textbf{Clean Accuracy} & \textbf{SN Attack Accuracy}\\ \midrule
MNIST Ensemble                    & 100.000\%      & 90.519\%           \\
MNIST Ensemble with NL            & 99.867\%       & 85.970\%           \\
MNIST Ensemble with NL+RV(0.05)   & 100.00\%       & 95.563\%           \\
CIFAR10 Ensemble                  & 93.333\%       & 89.304\%           \\
CIFAR10 Ensemble with NL          & 90.800\%       & 88.030\%           \\
CIFAR10 Ensemble with NL+RV(0.05) & 99.819\%       & 96.643\%          \\
 \bottomrule
\end{tabular}%
}
\label{tab:clean_vs_adversarial_single}
\end{subtable}
 
\vspace{0.25cm}
\begin{subtable}{1.0\linewidth} 
\centering
\subcaption{Adversarial classifications}
\vspace{-0.2cm}
    \resizebox{0.95\textwidth}{!}{%
\begin{tabular}{@{}llll@{}}
\toprule
\textbf{Model}           & \textbf{Correct} & \textbf{Target} & \textbf{Other} \\ \midrule
MNIST Ensemble           & 90.519\%         & 3.022\%         & 6.459\%        \\
MNIST Ensemble with NL   & 85.970\%         & 5.704\%         & 8.326\%        \\
CIFAR10 Ensemble         & 89.304\%         & 1.363\%         & 9.333\%        \\
CIFAR10 Ensemble with NL & 88.030\%         & 1.659\%         & 10.311\%       \\ \bottomrule
\end{tabular}%
}
\label{tab:classification_adversarial_single}
\end{subtable}

\vspace{0.25cm}
\begin{subtable}{1.0\linewidth}
\centering
\subcaption{Average adversarial perturbations}
\vspace{-0.2cm}
    \resizebox{0.95\textwidth}{!}{%
\begin{tabular}{@{}llll@{}}
\toprule
\textbf{Model}           & \textbf{Correct} & \textbf{Target} & \textbf{Other} \\ \midrule
MNIST Ensemble           & 15.956\%         & 15.879\%        & 22.567\%       \\
MNIST Ensemble with NL   & 14.646\%         & 18.653\%        & 23.276\%       \\
CIFAR10 Ensemble         & 4.161\%          & 0.960\%         & 4.692\%        \\
CIFAR10 Ensemble with NL & 4.128\%          & 0.759\%         & 4.500\%        \\ \bottomrule
\end{tabular}%
    }  
\label{tab:perturbation_adversarial_single}
\end{subtable}

     \caption{Distributions for Single Network adversarial inputs. (a) clean accuracy vs. attack accuracy; (b) breakdown of classifications; (c) average perturbations corresponding to classifications.} 
\label{tab:distribution_adversarial_single}
    \end{table}

\subsection{Random Single Network Attack}
\label{SN_results}
In this section, we look at the possible outcomes of an adversarial example crafted to defeat a single network, to see how it can potentially transfer across the ensemble. Since an adversarial example crafted for one network can potentially fool a different network \cite{papernot2016transferability}, we expect transferability in our ensemble of networks especially given that they have very similar model architectures. Given some sample input $s$ and target label $t$, for each $F^l$ we craft an adversarial example $A(F^l,s,t)$ using the Carlini-Wagner attack, and look at: 1) how each $F^{l'}$ classifies $A(F^l,s,t)$; 2) how the ensemble classifies the example through voting with and without Noisy Logit; and 3) how Rank Verification (with 5\% significance level) can further improve accuracy. 

We generate a set of 15 input samples, $s_k, k=1,...,15$. For each $s_k$ we craft an adversarial example $A(F^l,s_k,t_j)$ on network $F^l, l=1,...,50$, for targets $t_j, j=1,...,9$, for a total of $9\times50\times15=6750$ adversarial examples. We define the perturbation of an adversarial example as its normed difference with the original input over the norm of the original input, as follows:
{\small
\begin{equation}
\label{eqn:perturbation}
p(a;s) = \frac{||a-s||}{||s||},
\end{equation}
}
where $a$ is an adversarial example on the input $s$. We bucket the range of perturbations into 40 equally spaced bins, $x_b, b=1,...,40$. In the plots in this section, we aggregate by the perturbation bins and represent the bins by their mid-points on the x-axis.

\begin{figure}[t]
\centering
\begin{multicols}{2}
\begin{subfigure}{0.24\textwidth}
\subcaption{\label{fig:count_changed_target_cifar} To Target }
\vspace {-.2cm}
\begin{tikzpicture}
\begin{axis}[
    xlabel={Perturbation},
    ylabel={Count},
	xlabel style={at={(0.4,1.5ex)}},
    ylabel style={at={(0.18,10ex)}},
    xmin=-0.002, xmax=0.12,
    ymin=-.08, ymax=3.8,
    xtick={0, .03, .06, .09, .12},
    ytick={0,0.5, 1.0, 1.5, 2.0, 2.5, 3.0, 3.5},
    xticklabel style={/pgf/number format/fixed},
    mark size=0.7pt,
    ymajorgrids=true,
    grid style=dashed,
    legend style={font=\fontsize{5}{3}\selectfont},
    legend pos=north east,
    legend cell align={left},
    label style={font=\fontsize{7}{3}\selectfont},
    tick label style={font=\fontsize{4}{3}\selectfont},
    width=\textwidth,
    height=\textwidth,   
]
 
\addplot[
only marks,
    color=blue,
    mark=star,
    ]
file [skip first] {Plots/cifar_normal/Figure_flip2target_count.png_1.txt};
    
\addplot[
only marks,
    color=green,
    mark=o,
    only marks,
    ]
file [skip first] {Plots/cifar_normal/Figure_flip2target_count.png_0.txt};

\end{axis}
\end{tikzpicture}
\end{subfigure}

\columnbreak

\begin{subfigure}{0.24\textwidth}
\subcaption{\label{fig:count_changed_other_cifar}To Others}
\vspace {-.2cm}
\begin{tikzpicture}
\begin{axis}[
    xlabel={Perturbation},
    ylabel={Count},
   xlabel style={at={(0.4,1.5ex)}},
    ylabel style={at={(0.18,10ex)}},
    xmin=-0.002, xmax=0.122,
    ymin=-1, ymax=27,
    xtick={0, .03, .06, .09, .12},
    ytick={0,5, 10, 15, 20, 25},
    xticklabel style={/pgf/number format/fixed},
    mark size=0.7pt,
    ymajorgrids=true,
    grid style=dashed,
    legend style={font=\fontsize{5}{3}\selectfont},
	legend pos=north east,
    legend cell align={left},
    label style={font=\fontsize{7}{3}\selectfont},
    tick label style={font=\fontsize{4}{3}\selectfont},
    width=\textwidth,
    height=\textwidth,   
]
 
\addplot[
    color=blue,
    mark=star,
    only marks,
    ]
file [skip first] {Plots/cifar_normal/Figure_flip2other_count.png_1.txt};
    \legend{Noisy Adv. Inputs}
    
\addplot[
    color=green,
    mark=o,
    only marks,
    ]
file [skip first] {Plots/cifar_normal/Figure_flip2other_count.png_0.txt};
    \addlegendentry{Adv. Inputs}
    
\end{axis}
\end{tikzpicture}
\end{subfigure}
\end{multicols}
\vspace {-.6cm}
\caption{\label{fig:changed_cifar_single} CIFAR10: Counts of Single Networks Changed}
\end{figure}
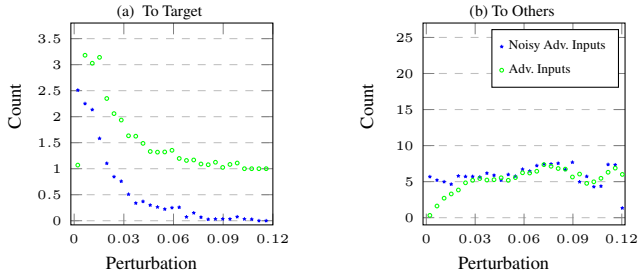

In Fig. \ref{fig:count_changed_target_mnist} and  \ref{fig:count_changed_target_cifar}, we have the average counts of networks whose classifications change to the target of the adversarial example, from some other original classification, i.e., for bucket $x_b$, the values $y_b$ on the y-axis are:
{\small
\begin{equation}\label{eqn:count_changed_target}
\begin{aligned}
y_b = \frac{1}{|x_b|}\sum_{x \in x_b} |{} & \{F^l:F^l(a)\neq F^l(s),F^l(a)=t,\\
& p(s,a)=x,l=1,...,50\}|.
\end{aligned}
\end{equation}
}
In Fig. \ref{fig:count_changed_other_mnist} and \ref{fig:count_changed_other_cifar}, the counts are on the classifications that change to something other than the target, or:
{\small
\begin{equation}
\label{eqn:count_changed_other}
\begin{aligned}
y_b = \frac{1}{|x_b|}\sum_{x \in x_b} |{} & \{F^l:F^l(a)\neq F^l(s),F^l(a)\neq t,\\
& p(s,a)=x,l=1,...,50\}|.
\end{aligned}
\end{equation}
}
These counts are averaged by perturbation bin. The green and blue curves represent the results corresponding to adversarial examples crafted on networks without and with Noisy Logit applied, respectively.

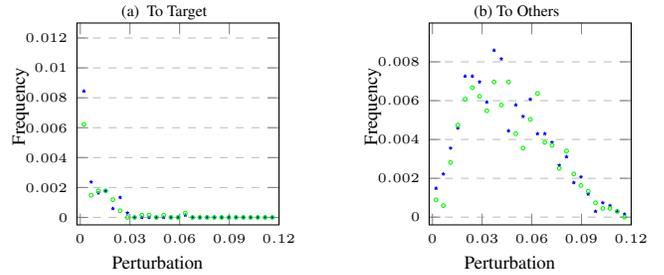
\begin{figure}[t]
\centering
\begin{multicols}{2}
\begin{subfigure}{0.24\textwidth}
\subcaption{\label{fig:freq_changed_target_cifar} To Target }
\vspace {-.2cm}
\begin{tikzpicture}
\begin{axis}[
    xlabel={Perturbation},
    ylabel={Frequency},
   xlabel style={at={(0.4,1.5ex)}},
    ylabel style={at={(0.18,10ex)}},
    xmin=-0.002, xmax=0.12,
    ymin=-0.0005, ymax=0.013,
    xtick={0, .03, .06, .09, .12},
    ytick={0.0, 0.002, 0.004, 0.006, 0.008, 0.01, 0.012},
    xticklabel style={/pgf/number format/fixed},
    yticklabel style={/pgf/number format/fixed, /pgf/number format/precision=3},
    scaled y ticks=false,
    mark size=0.7pt,
    ymajorgrids=true,
    grid style=dashed,
    legend style={font=\fontsize{5}{3}\selectfont},
    legend pos=north east,
    legend cell align={left},
    label style={font=\fontsize{7}{3}\selectfont},
    tick label style={font=\fontsize{4}{3}\selectfont},
    width=\textwidth,
    height=\textwidth,   
]
 
\addplot[
only marks,
    color=blue,
    mark=star,
    ]
file [skip first] {Plots/cifar_normal/Figure_adv2target_freq.png_1.txt};
    
\addplot[
only marks,
    color=green,
    mark=o,
    only marks,
    ]
file [skip first] {Plots/cifar_normal/Figure_adv2target_freq.png_0.txt};

\end{axis}
\end{tikzpicture}
\end{subfigure}

\columnbreak

\begin{subfigure}{0.24\textwidth}
\subcaption{\label{fig:freq_changed_other_cifar}To Others}
\vspace {-.2cm}
\begin{tikzpicture}
\begin{axis}[
    xlabel={Perturbation},
    ylabel={Frequency},
   xlabel style={at={(0.4,1.5ex)}},
    ylabel style={at={(0.18,10ex)}},
    xmin=-0.002, xmax=0.12,
    ymin=-.0004, ymax=0.01,
    xtick={0, .03, .06, .09, .12},
    ytick={0.0, 0.002, 0.004, 0.006, 0.008},
    xticklabel style={/pgf/number format/fixed},
    yticklabel style={/pgf/number format/fixed, /pgf/number format/precision=3},
    scaled y ticks=false,
    mark size=0.7pt,
    ymajorgrids=true,
    grid style=dashed,
    legend style={font=\fontsize{5}{3}\selectfont},
    legend cell align={left},
    label style={font=\fontsize{7}{3}\selectfont},
    tick label style={font=\fontsize{4}{3}\selectfont},
    width=\textwidth,
    height=\textwidth,   
]
 
\addplot[
    color=blue,
    mark=star,
    only marks,
    ]
file [skip first] {Plots/cifar_normal/Figure_adv2other_freq.png_1.txt};
    
\addplot[
    color=green,
    mark=o,
    only marks,
    ]
file [skip first] {Plots/cifar_normal/Figure_adv2other_freq.png_0.txt};
    
\end{axis}
\end{tikzpicture}
\end{subfigure}

\end{multicols}

%
\vspace {-.6cm}
\caption{\label{fig:changed_cifar_ensemble} CIFAR10: Ensemble Aggregate Outputs Changed}

\end{figure}

\begin{figure}[t]
\centering
\begin{multicols}{2}
\begin{subfigure}{0.24\textwidth}
\subcaption{\label{fig:avg_accuracy_mnist} MNIST}
\vspace {-.2cm}
\begin{tikzpicture}
\begin{axis}[
    xlabel={Perturbation},
    ylabel={Average Accuracy},
   xlabel style={at={(0.44,1.5ex)}},
    ylabel style={at={(0.25,10ex)}},
    xmin=-0.015, xmax=0.5,
    ymin=-.05, ymax=1.2,
    xtick={0.0, 0.1, 0.2,0.3,0.4,0.5},
    ytick={0,0.2,0.4,0.6,0.8,1.0},
    mark size=0.7pt,
    ymajorgrids=true,
    grid style=dashed,
    legend style={font=\fontsize{5}{3}\selectfont},
    legend pos=south west,
    legend cell align={left},
    label style={font=\fontsize{7}{3}\selectfont},
    tick label style={font=\fontsize{4}{3}\selectfont},
    width=\textwidth,
    height=\textwidth,   
]

\addplot[
only marks,
    color=blue,
    mark=o,
    ]
file [skip first] {Plots/mnist_normal/trans_avg_accuracy.png_1.txt};
	
\addplot[
only marks,
    color=green,
    mark=star,
    ]
file [skip first] {Plots/mnist_normal/trans_avg_accuracy.png_0.txt};

\end{axis}
\end{tikzpicture}
\end{subfigure}

\columnbreak

\begin{subfigure}{0.24\textwidth}
\caption{\label{fig:avg_accuracy_cifar} CIFAR10}
\vspace {-.2cm}
\begin{tikzpicture}
\begin{axis}[
    xlabel={Perturbation},
    ylabel={Average Accuracy},
   xlabel style={at={(0.44,1.5ex)}},
    ylabel style={at={(0.25,10ex)}},
    xmin=-0.002, xmax=0.118,
    ymin=0.2, ymax=1.1,
    xtick={0, .03, .06, .09, .12},
    ytick={0,0.3,0.4,0.5,0.6,0.7, 0.8, 0.9, 1.0},
    xticklabel style={/pgf/number format/fixed},
    mark size=0.7pt,
    ymajorgrids=true,
    grid style=dashed,
    legend style={font=\fontsize{5}{3}\selectfont},
    legend pos=south west,
    legend cell align={left},
    label style={font=\fontsize{7}{3}\selectfont},
    tick label style={font=\fontsize{4}{3}\selectfont},
    width=\textwidth,
    height=\textwidth,   
]
 
\addplot[
only marks,
    color=blue,
    mark=star,
    ]
file [skip first] {Plots/cifar_normal/trans_avg_accuracy.png_1.txt};
    \legend{Noisy Adv. Inputs}
    
\addplot[
only marks,
    color=green,
    mark=o,
    ]
file [skip first] {Plots/cifar_normal/trans_avg_accuracy.png_0.txt};
    \addlegendentry{Adv. Inputs}    

\end{axis}
\end{tikzpicture}
\end{subfigure}
\end{multicols}
\vspace {-.6cm}
\caption{\label{fig:perturbation_accuracy} Average accuracy vs. perturbation}
\end{figure}
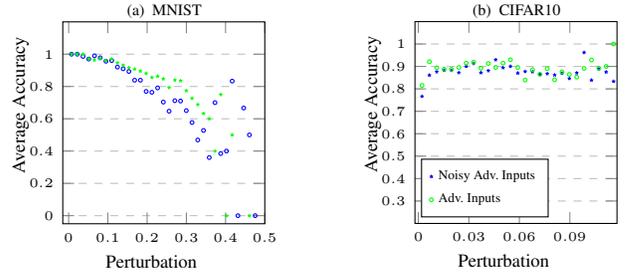

In Fig. \ref{fig:freq_changed_target_mnist} and \ref{fig:freq_changed_target_cifar} we show the frequencies of aggregate outputs of the ensemble which changed to the target of the adversarial example, i.e.:
{\small
\begin{equation}
\label{eqn:freq_changed_target}
\begin{aligned}
y_b = \frac{1}{N}\sum_{x \in x_b} |{} & \{F^*(a):F^*(a)\neq F^*(s),F^*(a)=t,\\
& p(s,a)=x\}|,
\end{aligned}
\end{equation}
}
where $F^*(\cdot)$ represents the aggregate output by voting among the ensemble. Similarly, we have in Fig. \ref{fig:freq_changed_other_mnist} and \ref{fig:freq_changed_other_cifar}the frequencies corresponding to some label other than the target:
{\small
\begin{equation}
\label{eqn:freq_changed_other}
\begin{aligned}
y_b = \frac{1}{N}\sum_{x \in x_b} |{} & \{F^*(a):F^*(a)\neq F^*(s),F^*(a)\neq t,\\
& p(s,a)=x\}|.
\end{aligned}
\end{equation}
}
The frequencies are obtained by normalizing the total changed outputs by the total number of adversarial examples, which is $N=6750$ in this case.
In Fig. \ref{fig:perturbation_accuracy} we show the average accuracy of the ensemble by perturbation bin. From these plots, we observe 1) applying Noisy Logit clearly reduces transferability rate for CIFAR10 networks, where in Fig. \ref{fig:count_changed_target_cifar} we notice a decrease in the number of networks in the ensemble whose classifications change to target when Noisy Logit is applied, and consistently so across different perturbation bins; 2) applying Noisy Logit changes the distribution of perturbations for MNIST, where higher frequencies of larger perturbations are observed.


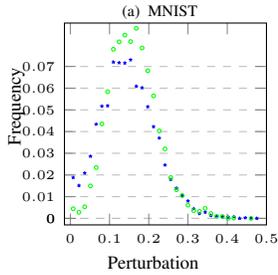
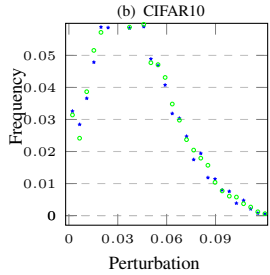
\begin{figure}[t]
\centering
\begin{multicols}{2}
\begin{subfigure}{0.24\textwidth}
\caption{\label{fig:perturb_distribution_correct_mnist} MNIST}
\vspace {-.2cm}
\begin{tikzpicture}
\begin{axis}[
    xlabel={Perturbation},
    ylabel={Frequency},
   xlabel style={at={(0.44,1.5ex)}},
    ylabel style={at={(0.22,10ex)}},
    xmin=-0.015, xmax=0.5,
    ymin=-.003, ymax=.09,
    xtick={0.0, 0.1, 0.2,0.3,0.4,0.5},
    ytick={0,0, 0.01, 0.020,0.03,0.04,0.05,0.06,0.07},
    yticklabel style={/pgf/number format/fixed, /pgf/number format/precision=2},
    scaled y ticks=false,
    mark size=0.7pt,
    ymajorgrids=true,
    grid style=dashed,
    legend style={font=\fontsize{5}{3}\selectfont},
    legend pos=south east,
    legend cell align={left},
    label style={font=\fontsize{7}{3}\selectfont},
    tick label style={font=\fontsize{4}{3}\selectfont},
    width=\textwidth,
    height=\textwidth,   
]
 
\addplot[
    color=blue,
    mark=star,
    only marks,
    ]
file [skip first] {Plots/mnist_normal/Figure_adv2correct_freq.png_1.txt};
    
\addplot[
    color=green,
    mark=o,
    only marks,
    ]
file [skip first] {Plots/mnist_normal/Figure_adv2correct_freq.png_0.txt};
    
\end{axis}
\end{tikzpicture}
\end{subfigure}

\columnbreak

\begin{subfigure}{0.24\textwidth}
\caption{\label{fig:perturb_distribution_correct_cifar} CIFAR10}
\vspace {-.2cm}
\begin{tikzpicture}
\begin{axis}[
    xlabel={Perturbation},
    ylabel={Frequency},
   xlabel style={at={(0.44,1.5ex)}},
    ylabel style={at={(0.22,10ex)}},
    xmin=-0.002, xmax=0.122,
    ymin=-.003, ymax=.06,
    xtick={0, .03, .06, .09, .13},
    ytick={0,0, 0.01, 0.020,0.03,0.04,0.05},
    xticklabel style={/pgf/number format/fixed},
    yticklabel style={/pgf/number format/fixed, /pgf/number format/precision=2},
    scaled y ticks=false,
    mark size=0.7pt,
    ymajorgrids=true,
    grid style=dashed,
    legend style={font=\fontsize{5}{3}\selectfont},
    label style={font=\fontsize{7}{3}\selectfont},
    tick label style={font=\fontsize{4}{3}\selectfont},
    width=\textwidth,
    height=\textwidth,   
]
 
\addplot[
    color=blue,
    mark=star,
    only marks,
    ]
file [skip first] {Plots/cifar_normal/Figure_adv2correct_freq.png_1.txt};
    
\addplot[
    color=green,
    mark=o,
    only marks,
    ]
file [skip first] {Plots/cifar_normal/Figure_adv2correct_freq.png_0.txt};
    
\end{axis}
\end{tikzpicture}
\end{subfigure}

\end{multicols}

%
\vspace {-.6cm}
\caption{\label{fig:perturbation_distribution_mnistcifar} Perturbation distribution of correct outputs}

\end{figure}
We summarize the results in Table \ref{tab:distribution_adversarial_single}. Observe that for CIFAR10 Ensemble with Noisy Logit, nearly identical accuracy to the original clean accuracy is achieved for adversarial examples that target any single network in the ensemble. We calculate accuracy as the ratio of the number of correct aggregate outputs over the total number (6750) of test attacks. 

\begin{figure}[t]
\centering
\begin{multicols}{2}
\begin{subfigure}{0.24\textwidth}
\subcaption{\label{fig:guarantee_SRN_mnist} MNIST SRN }
\vspace {-.2cm}
\begin{tikzpicture}
\begin{axis}[
    xlabel={Perturbation},
    ylabel={Count},
   xlabel style={at={(0.4,1.5ex)}},
    ylabel style={at={(0.18,10ex)}},
    xmin=0, xmax=15,
    ymin=0, ymax=5,
    xtick={0,3,6,9,12,15},
    ytick={0,0.5,1,1.5,2,2.5,3,3.5},
    mark size=0.9pt,
    ymajorgrids=true,
    grid style=dashed,
    legend style={font=\fontsize{5}{3}\selectfont},
    legend pos=north west,
    legend cell align={left},
    label style={font=\fontsize{7}{3}\selectfont},
    tick label style={font=\fontsize{4}{3}\selectfont},
    width=\textwidth,
    height=\textwidth,   
]
 
\addplot[
only marks,
    color=blue,
    mark=diamond,
    ]
file [skip first] {Plots/mnist_normal/Figure_distortmin_SRN.txt};
    \legend{Min Distortion}
    
\addplot[
only marks,
    color=green,
    mark=-,
	very thick,
	dashed,
    ]
file [skip first] {Plots/mnist_normal/Figure_cert05_SRN.txt};
    \addlegendentry{Guarantee}


\end{axis}
\end{tikzpicture}
\end{subfigure}
\columnbreak

\begin{subfigure}{0.24\textwidth}
\subcaption{\label{fig:guarantee_sup2_mnist} MNIST SI2 }
\vspace {-.2cm}
\begin{tikzpicture}
\begin{axis}[
    xlabel={Perturbation},
    ylabel={Count},
   xlabel style={at={(0.4,1.5ex)}},
    ylabel style={at={(0.18,10ex)}},
    xmin=0, xmax=30,
    ymin=0, ymax=8,
    xtick={0,5,10,15,20,25,30},
    ytick={0,1,2,3,4,5,6},
    mark size=0.9pt,
    ymajorgrids=true,
    grid style=dashed,
    legend style={font=\fontsize{5}{3}\selectfont},
    legend pos=north west,
    legend cell align={left},
    label style={font=\fontsize{7}{3}\selectfont},
    tick label style={font=\fontsize{4}{3}\selectfont},
    width=\textwidth,
    height=\textwidth,   
]
 
\addplot[
only marks,
    color=blue,
    mark=diamond,
    ]
file [skip first] {Plots/mnist_normal/Figure_distortmin_sup2.txt};
    
\addplot[
only marks,
    color=green,
    mark=-,
	very thick,
	dashed,
    ]
file [skip first] {Plots/mnist_normal/Figure_cert05_SRN.txt};


\end{axis}
\end{tikzpicture}
\end{subfigure}

\end{multicols}
\vspace {-.6cm}
\caption{\label{fig:guarantee_mnist} MNIST: Robustness $R$ vs. Attack $R$}
\end{figure}
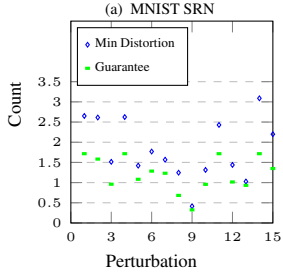
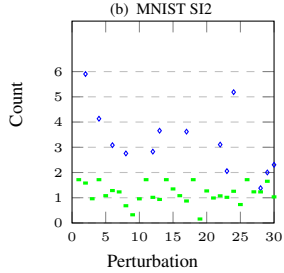

\begin{figure}[t]
\centering
\begin{multicols}{2}
\begin{subfigure}{0.24\textwidth}
\subcaption{\label{fig:guarantee_SRN_cifar} CIFAR SRN }
\vspace {-.2cm}
\begin{tikzpicture}
\begin{axis}[
    xlabel={Input Sample},
    ylabel={Radius},
   xlabel style={at={(0.4,1.5ex)}},
    ylabel style={at={(0.18,10ex)}},
    xmin=0, xmax=15,
    ymin=0, ymax=1,
    xtick={0,3,6,9,12,15},
    ytick={0,0.1,0.2,0.3,0.4,0.5,0.6,0.7,0.8},
    mark size=0.9pt,
    ymajorgrids=true,
    grid style=dashed,
    legend style={font=\fontsize{5}{3}\selectfont},
    legend pos=north west,
    legend cell align={left},
    label style={font=\fontsize{7}{3}\selectfont},
    tick label style={font=\fontsize{4}{3}\selectfont},
    width=\textwidth,
    height=\textwidth,   
]
 
\addplot[
only marks,
    color=blue,
    mark=diamond,
    ]
file [skip first] {Plots/cifar_normal/Figure_distortmin_SRN_changed.txt};
    \legend{Min Distortion}
    
\addplot[
only marks,
    color=green,
    mark=-,
	very thick,
	dashed,
    ]
file [skip first] {Plots/cifar_normal/Figure_cert003_SRN.txt};
    \addlegendentry{Guarantee}

\addplot[
only marks,
    color=yellow,
    mark=x,
    ]
file [skip first] {Plots/cifar_normal/Figure_cert003_vr.txt};

\addplot[
only marks,
    color=red,
    mark=x,
    ]
file [skip first] {Plots/cifar_normal/Figure_cert003_abstain.txt};

\end{axis}
\end{tikzpicture}
\end{subfigure}
\columnbreak

\begin{subfigure}{0.24\textwidth}
\subcaption{\label{fig:guarantee_sup2_cifar} CIFAR SI2 }
\vspace {-.2cm}
\begin{tikzpicture}
\begin{axis}[
    xlabel={Input Sample},
    ylabel={Radius},
   xlabel style={at={(0.4,1.5ex)}},
    ylabel style={at={(0.18,10ex)}},
    xmin=0, xmax=30,
    ymin=0, ymax=1,
    xtick={0,5,10,15,20,25,30},
    ytick={0,0.2,0.4,0.6,0.8},
    mark size=0.9pt,
    ymajorgrids=true,
    grid style=dashed,
    legend style={font=\fontsize{5}{3}\selectfont},
    legend pos=north west,
    legend cell align={left},
    label style={font=\fontsize{7}{3}\selectfont},
    tick label style={font=\fontsize{4}{3}\selectfont},
    width=\textwidth,
    height=\textwidth,   
]
 
\addplot[
only marks,
    color=blue,
    mark=diamond,
    ]
file [skip first] {Plots/cifar_normal/Figure_distortmin_sup2_changed.txt};
    
\addplot[
only marks,
    color=green,
    mark=-,
	very thick,
	dashed,
    ]
file [skip first] {Plots/cifar_normal/Figure_cert003_SRN.txt};

\addplot[
only marks,
    color=yellow,
    mark=x,
    ]
file [skip first] {Plots/cifar_normal/Figure_cert003_vr.txt};

\addplot[
only marks,
    color=red,
    mark=x,
    ]
file [skip first] {Plots/cifar_normal/Figure_cert003_abstain.txt};

\end{axis}
\end{tikzpicture}
\end{subfigure}

\end{multicols}
\vspace {-.6cm}
\caption{\label{fig:guarantee_cifar} CIFAR: Robustness $R$ vs. Attack $R$}
\end{figure}
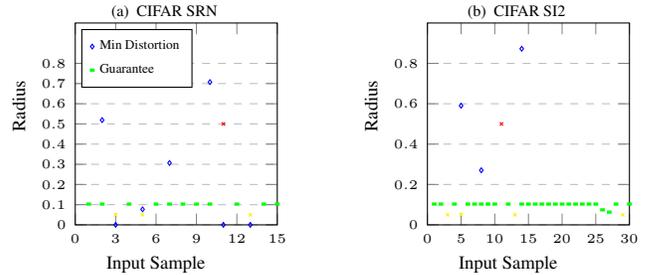

\subsection{Superimposition Attacks}
\label{SI_results}
In this section we consider superimposition attacks consisting of adversarial examples targeting two or three of the networks in the ensemble. Due to the large number of possible subsets of size two or three, we do not consider every such combination; instead, since the objective of crafting an adversarial example is to minimize the perturbations while causing a network to incorrectly classify, we consider a greedy type superimposition where adversarial examples of minimal perturbations are used. The setup is in Table \ref{tab:si_test_config}.

 \begin{table}[] 
 \resizebox{0.45\textwidth}{!}{%
\begin{tabular}{@{}llll@{}}
\toprule
\textbf{Attack} & \textbf{\# sample inputs} & \textbf{\# adversarial examples crafted} & \textbf{\# tests} \\ \midrule
SI2             & 30                        & 30x9x50=13500                            & 180               \\
SI3             & 20                        & 20x9x50=9000                             & 270               \\ \bottomrule
\end{tabular}
}
\caption{Setup for testing superimposition attacks.} 
\label{tab:si_test_config}
\end{table}
\vspace{.2cm}
\noindent
\textbf{Superimposition of Two Adversarial Examples}
\noindent

In Table \ref{tab:accuracy_clean_vs_super2_SN} we have the average accuracy over individual networks, where in the left column the accuracy is on the original images, in the right column the accuracy is on the adversarial examples. We see that with a superimposition of only two images, the average accuracy of a single MNIST network reduced to 55.30\% from 96.80\%; whereas for a CIFAR10 network the average accuracy is reduced from 79.53\% to 66.48\%, a much smaller reduction. This again suggests that CIFAR10 networks are more robust to transferability. Observe that with Noisy Logit, the accuracy reduction is less prominent.

 \begin{table}[h]
 \centering
 \begin{subtable}{1.0\linewidth} 
\centering
\subcaption{Average single network clean accuracy vs. attack accuracy}
\vspace{-0.2cm} 
    \resizebox{0.95\textwidth}{!}{%
\begin{tabular}{@{}lll@{}}
\toprule
\textbf{Network}         & \textbf{Clean Accuracy} & \textbf{SI2 Attack Accuracy} \\ \midrule
MNIST Single           & 96.800\%                & 55.296\%                    \\
MNIST Single with NL   & 90.044\%                & 73.007\%                    \\
CIFAR10 Single         & 79.533\%                & 66.481\%                    \\
CIFAR10 Single with NL & 76.570\%                & 72.259\%             \\ \bottomrule      
\end{tabular}
    }
\label{tab:accuracy_clean_vs_super2_SN}
    \end{subtable}    
 
\vspace{0.25cm}
  \begin{subtable}{1.0\linewidth} 
\centering
\subcaption{Accuracy  of the ensembles, with Noisy Logit (NL), and with NL + Rank Verification at 5\% significance level (NL+RV(0.05))}
\vspace{-0.2cm} 
    \resizebox{0.95\textwidth}{!}{%
\begin{tabular}{@{}lll@{}}
\toprule
\textbf{Model}                    & \textbf{Clean Accuracy} & \textbf{SI2 Attack Accuracy} \\ \midrule
MNIST Ensemble                    & 100.000\%               & 66.296\%                    \\
MNIST Ensemble with NL            & 100.000\%               & 94.074\%                    \\
MNIST Ensemble with NL+RV(0.05)   & 100.000\%               & 99.099\%                    \\
CIFAR10 Ensemble                  & 90.000\%                & 85.926\%                    \\
CIFAR10 Ensemble with NL          & 87.778\%                & 87.037\%                    \\
CIFAR10 Ensemble with NL+RV(0.05) & 92.035\%                & 91.628\%                    \\ \bottomrule
\end{tabular}%
    }
\label{tab:accuracy_clean_vs_super2}
    \end{subtable}
    
\vspace{0.25cm}
\begin{subtable}{1.0\linewidth} 
\centering
\subcaption{Breakdown of classifications}
\vspace{-0.2cm}
    \resizebox{0.95\textwidth}{!}{%
\begin{tabular}{@{}llll@{}}
\toprule
\textbf{Model}           & \textbf{Correct} & \textbf{Target} &  \textbf{Other} \\ \midrule
MNIST Ensemble           & 66.296\%         & 13.333\%        & 20.370\% \\
MNIST Ensemble with NL   & 94.074\%         & 0.370\%         & 5.556\%  \\
CIFAR10 Ensemble         & 85.926\%         & 1.111\%         & 12.963\% \\
CIFAR10 Ensemble with NL & 87.037\%         & 1.481\%         & 11.481\% \\ \bottomrule
\end{tabular}
    }
\label{tab:classification_clean_vs_super2}
    \end{subtable}

\vspace{0.25cm}
\begin{subtable}{1.0\linewidth}
\centering
\subcaption{Average perturbations corresponding to classifications}
\vspace{-0.2cm}
    \resizebox{0.95\textwidth}{!}{%
\begin{tabular}{@{}llll@{}}
\toprule
\textbf{Model}           & \textbf{Correct} & \textbf{Target} & \textbf{Other} \\ \midrule
MNIST Ensemble           & 14.628\%         & 25.713\%        & 24.323\%       \\
MNIST Ensemble with NL   & 9.564\%          & 23.179\%        & 24.446\%       \\
CIFAR10 Ensemble         & 3.685\%          & 0.000\%         & 4.782\%        \\
CIFAR10 Ensemble with NL & 2.651\%          & 0.000\%         & 3.385\%        \\ \bottomrule
\end{tabular}
    }
     \label{tab:perturb_clean_vs_super2}
    \end{subtable}
    \caption{Distributions for Superimposition ($2\times$) of adversarial inputs. } 

\label{tab:distrib_clean_vs_super2}
    \end{table}
Observe in Fig. \ref{tab:accuracy_clean_vs_super2} how much each model improves upon the single network case in terms of clean and adversarial accuracy. Moreover, when both Noisy Logit and Rank Verification (at 5\% significance level) are applied, the model is able to achieve superior adversarial accuracy, one that's even higher than the single network clean accuracy.

We provide the distribution of perturbations corresponding to Table \ref{tab:classification_clean_vs_super2} in Table \ref{tab:perturb_clean_vs_super2}. Observe that the average perturbation for MNIST Ensemble with Noisy Logit is smaller than that without Noisy Logit, consistent with Fig. \ref{fig:perturb_distribution_correct_mnist}, where applying noise shifts the distribution of perturbations toward the tails.

\vspace{.2cm}
\noindent
\textbf{Superimposition of Three Adversarial Examples}\\
\noindent
For the superimposition of three adversarial examples (Table~\ref{tab:distribution_clean_vs_super3}),  we observe similar results as two adversarial examples. We note that under this attack, a simple ensemble performs no better than a single network, with only $22\%$ accuracy. However, with Noisy Logit and Rank Verification (at 5\% significance level) applied, the model still achieves great accuracy, especially for CIFAR10.
%

    \begin{table}[h]
 \centering
\begin{subtable}{1.0\linewidth} 
\centering
\subcaption{Average single network clean accuracy vs. attack accuracy}
\vspace{-0.2cm} 
    \resizebox{0.95\textwidth}{!}{%
\begin{tabular}{@{}lll@{}}
\toprule
\textbf{Network}         & \textbf{Clean Accuracy} & \textbf{SI3 Attack Accuracy} \\ \midrule
MNIST Single           & 95.500\%                & 21.900\%                    \\
MNIST Single with NL   & 88.878\%                & 54.267\%                    \\
CIFAR10 Single         & 84.800\%                & 69.289\%                    \\
CIFAR10 Single with NL & 81.578\%                & 71.089\%                    \\ \bottomrule
\end{tabular}
    }  
\label{tab:accuracy_avg_clean_vs_super3}
    \end{subtable}  

\vspace{0.25cm} 
\begin{subtable}{1.0\linewidth} 
\centering
\subcaption{Accuracy of the ensembles, with Noisy Logit (NL), and with NL + Rank Verification at 5\% significance level (NL+RV(0.05))}
\vspace{-0.2cm} 
    \resizebox{0.95\textwidth}{!}{%
\begin{tabular}{@{}lll@{}}
\toprule
\textbf{Model}                    & \textbf{Clean Accuracy} & \textbf{SI3 Attack   Accuracy} \\ \midrule
MNIST Ensemble                    & 100.000\%               & 22.222\%                      \\
MNIST Ensemble with NL            & 100.000\%               & 72.222\%                      \\
MNIST Ensemble with NL+RV(0.05)   & 100.000\%               & 83.471\%                      \\
CIFAR10 Ensemble                  & 95.000\%                & 88.889\%                      \\
CIFAR10 Ensemble with NL          & 92.778\%                & 88.333\%                      \\
CIFAR10 Ensemble with NL+RV(0.05) & 100.000\%               & 98.485\%                      \\ \bottomrule
\end{tabular}
    }  
\label{tab:accuracy_clean_vs_super3}
    \end{subtable} 

\vspace{0.25cm} 
\begin{subtable}{1.0\linewidth} 
\centering
\subcaption{Breakdown of classifications}
\vspace{-0.2cm}  
    \resizebox{0.95\textwidth}{!}{%
\begin{tabular}{@{}llll@{}}
\toprule
\textbf{Model}           & \textbf{Correct} & \textbf{Target} & \textbf{Other} \\ \midrule
MNIST Ensemble           & 22.222\%         & 59.444\%        & 18.333\%       \\
MNIST Ensemble with NL   & 72.222\%         & 10.556\%        & 17.222\%       \\
CIFAR10 Ensemble         & 88.889\%         & 0.556\%         & 10.556\%       \\
CIFAR10 Ensemble with NL & 88.333\%         & 1.111\%         & 10.556\%       \\ \bottomrule
\end{tabular}
        }    
    \label{tab:classification_clean_vs_super3}
        \end{subtable}

\vspace{0.25cm} 
\begin{subtable}{1.0\linewidth} 
\centering
\subcaption{Average perturbations corresponding to classifications}
\vspace{-0.2cm} 
    \resizebox{0.95\textwidth}{!}{%
\begin{tabular}{@{}llll@{}}
\toprule
\textbf{Model}           & \textbf{Correct} & \textbf{Target} & \textbf{Other} \\ \midrule
MNIST Ensemble           & 12.039\%         & 29.658\%        & 34.468\%       \\
MNIST Ensemble with NL   & 10.557\%         & 31.702\%        & 28.083\%       \\
CIFAR10 Ensemble         & 3.923\%          & 0.000\%         & 5.570\%        \\
CIFAR10 Ensemble with NL & 3.744\%          & 0.000\%         & 4.545\%        \\ \bottomrule
\end{tabular}
            }       
            \label{tab:perturb_clean_vs_super3}
            \end{subtable}
 
            \caption{Distributions for Superimposition ($3\times$) of adversarial inputs. } 
        \label{tab:distribution_clean_vs_super3}
            \end{table}       

\vspace {.2cm}
\noindent    
\textbf{Measuring Robustness}\\
We plot the robustness radius $R$ computed from our certification procedure, against the minimum computed $L^2$ norms of the adversarial perturbations that altered the original output, for the inputs used in experiments above. If the adversarial perturbation does not change the output for a given input sample, there is no minimum distortion radius for that sample. 

For samples which our procedure fails to certify (i.e. $\underline{p_A} \leq 0.5$), we mark it with a red x at 0.5. For samples that are certified but whose predictions fail Rank Verification with p-value $\geq 0.05$, we mark it with a yellow x at 0.05. We consider that the procedure should abstain from certification in both cases. We use $n=10000$ and $\alpha=0.05$ for both datasets. For MNIST we use $\sigma=0.5$ and for CIFAR10 $\sigma=0.03$ for certification.

Note that although in general a larger value of $\sigma$ leads to a larger value of certified robust $R$, it might also increase the likelihood that the procedure fails to certify. 
We demonstrate the certified radius $R$ for the Single Random Network (SRN) and Superimposition $2\times$ (SI2) attacks Fig.\ref{fig:guarantee_mnist} and \ref{fig:guarantee_cifar}. Observe that the certified radius is quite tight in some cases, where a slightly larger distortion already induces the model to change its prediction.
\section{Conclusion}
In this paper we introduced an approach to protect image classification networks from adversarial examples. The approach is composed of two mechanisms - Noisy Logit and Ensemble Voting, which were evaluated in Section \ref{testing_results}. We saw that Ensemble Voting improves accuracy over the base model, while Noisy Logit reduces transferability across different networks in classifying adversarial examples. Moreover, the approach combining the two mechanisms was shown to have comparable accuracy in classifying adversarial examples as in classifying genuine inputs for both MNIST and CIFAR10, where further improved accuracy was achieved with Rank Verification. Using Noisy Logit impedes the adversary's ability to accurately solve the optimization problem for crafting adversarial examples, as solving the problem requires access to outputs at some layers which have been tampered by the Noisy Logit mechanism. Ensemble Voting works on white-box attacks and is a mechanism that provides resilience as well as improves accuracy. Since using Noisy Logit reduces accuracy in general, the addition of Ensemble Voting complements the approach by improving the reduced accuracy.

There are a number of future directions to extend the current work. First, we can consider Ensemble Voting using a collection of networks with very different architectures (rather than the ones considered here which are the same up to the temperature constant). In the work of \cite{9156305}, it was demonstrated that some architectures are more robust to adversarial examples, we expect that our model would benefit from using more robust architectures since each individual network would be more robust and transferability could also be reduced as well. 
One can also study the relation between the number of networks to be used with the amount of noise to be added to the inputs, since it's not obvious how their relationship works in the robustness guarantee. It'd be interesting to research to perform similar experiments on another dataset as well, since MNIST and CIFAR10 exhibit different transferability properties.

\subsubsection*{Acknowledgements}
We thank anonymous reviewers for their suggestions and feedback. Supports from Vector Institute  and Natural Sciences \& Engineering Research Council of Canada (NSERC) are acknowledged. 
%
%
%

\bibliography{ensemble_main}

\end{document}